%% file: 00-main.tex
\documentclass[10pt,twocolumn,letterpaper]{article}

\usepackage{cvpr}
\usepackage{times}
\usepackage{epsfig}
\usepackage{graphicx}
\usepackage{amsmath}
\usepackage{amssymb}
\usepackage{microtype}
\usepackage{booktabs}
\usepackage{tabularx}
\usepackage{subfig}
\usepackage{xcolor}
\usepackage{bbding}
\usepackage{tabulary}
\usepackage{url}
\usepackage{multirow,overpic}
\newcolumntype{x}[1]{>{\centering\arraybackslash}p{#1pt}}

\newlength\savewidth\newcommand\shline{\noalign{\global\savewidth\arrayrulewidth
  \global\arrayrulewidth 1pt}\hline\noalign{\global\arrayrulewidth\savewidth}}

\usepackage[pagebackref=true,breaklinks=true,letterpaper=true,colorlinks,bookmarks=false]{hyperref}

\makeatletter
\newcommand{\printfnsymbol}[1]{%
  \textsuperscript{\@fnsymbol{#1}}%
}
\makeatother

\cvprfinalcopy 

\def\cvprPaperID{****} 

\ifcvprfinal\fi
\begin{document}

\title{Actor-Context-Actor Relation Network for Spatio-Temporal Action Localization\\
{\large 1st place solution for AVA-Kinetics Crossover in AcitivityNet Challenge 2020}}

\author{Siyu Chen$^2$\thanks{Equal contribution} \quad Junting Pan$^1$\printfnsymbol{1} \quad Guanglu Song$^2$ \quad Manyuan Zhang$^2$ \\
Hao Shao$^2$ \quad Ziyi Lin$^1$  \quad Jing Shao$^2$ \quad Hongsheng Li$^1$ \quad Yu Liu$^2$\\
$^1$CUHK-SenseTime Joint Lab, The Chinese University of Hong Kong\\
$^2$SenseTime Research
}

\maketitle

\begin{abstract}
   This technical report introduces our winning solution to the spatio-temporal action localization track, AVA-Kinetics Crossover, in ActivityNet Challenge 2020. Our entry is mainly based on Actor-Context-Actor Relation Network \cite{pan2020actorcontextactor}. We describe technical details for the new AVA-Kinetics dataset, together with some experimental results. Without any bells and whistles, we achieved \textbf{39.62 mAP} on the test set of AVA-Kinetics, which outperforms other entries by a large margin. Code will be available at: \href{https://github.com/Siyu-C/ACAR-Net}{https://github.com/Siyu-C/ACAR-Net}.
\end{abstract}


\input{01-Approach}
\input{02-Experiments}

{\small
\bibliographystyle{ieee_fullname}
\bibliography{egbib}
}
\end{document}

%% file: 01-Approach.tex
\section{Method}
\label{sec:Approach}

Our approach features Actor-Context-Actor Relation Network (ACAR-Net), details of which can be found in \cite{pan2020actorcontextactor}. Our proposed ACAR-Net gives an efficient yet effective algorithm to explicitly model and utilize higher-order relations built upon the basic first-order actor-context relations for assisting action localization.

\subsection{Overall Framework}
We first introduce our overall framework for action localization, where our proposed ACAR-Net is its key module for high-order relation modeling. The framework is designed to detect all persons in an input video clip and predict their action labels. 

We combine an \textit{off-the-shelf person detector} (\textit{e.g.} Faster R-CNN~\cite{ren2015faster}) with a \textit{video backbone network} (\textit{e.g.} I3D~\cite{carreira2017quo}). In details, the detector operates on the center frame (\textit{i.e.} key frame) of the clip and obtains $N$ detected actors. Such detected boxes are duplicated to other frames of the clip. In the mean time, the backbone network extracts a spatio-temporal feature volume from the input video clip. We perform average pooling along the temporal dimension considering computational efficiency, which results in a feature map $V \in \mathbb{R}^{C\times H\times W}$, where $C, H, W$ correspond to channel, height and width respectively. We apply RoIAlign~\cite{he2017mask} ($7\times 7$ spatial output) followed by spatial max pooling to the feature map $V$ and the $N$ actor boxes, producing a series of $N$ actor features, $A_1, A_2, \cdots, A_N \in \mathbb{R}^C$. Each actor feature describes the spatio-temporal appearance and motion of one Region of Interest (RoI).

The final classification head takes the aforementioned video feature map $V$ and RoI features $\{A_i\}_{i=1}^N$ as inputs, and outputs the final action predictions possibly after relation reasoning. The simplest baseline is a "Linear" head, which directly applies a linear classifier to the RoI features.

\subsection{Actor-Context-Actor Relation Network}
\label{sec:hro}

We first encode the first-order actor-context relations between each actor and each spatial location of the spatio-temporal context. More specifically, it concatenates each actor feature $A_i \in \mathbb{R}^{C}$ to all $H\times W$ spatial locations of the context feature $V \in \mathbb{R}^{C \times H \times W}$ to form a concatenated feature map $F'_i \in \mathbb{R}^{2C \times H \times W}$. The actor-context relation feature $F^{i}$ for actor $i$ can then be computed by applying convolutions to this concatenated feature map.

Based on the actor-context relations, we further add a \textbf{High-order Relation Reasoning Operator} (HR$^2$O) for modeling the connections established on first-order relations, which are indirect relations mostly ignored by previous methods. Let $F^{i}_{x,y}$ record the first-order feature between the actor $A_i$ and the scene context $V$ at the spatial location $(x,y)$. We introduce \textit{High-order Relation Reasoning}, in order to explicitly model the relations between first-order actor-context relations, which encode more informative scene semantics. However, since there are a large number of actor-context relation features, $F^i_{x,y}$, $i \in \{1,\cdots, N\}, x \in [1,H], y \in [1,W]$, the number of their possible pairwise combinations are generally overwhelming. We therefore propose to focus on learning the high-order relations between different actor-context relations at the same spatial location $(x,y)$, i.e.\ $F^i_{x,y}$ and $F^j_{x,y}$. In this way, the proposed relational reasoning operator limits the relation learning to second-order actor-context-actor relations, \textit{i.e.} two actors $i$ and $j$ can be associated via the same spatial context as $i \leftrightarrow (x,y) \leftrightarrow j$ to help the prediction of their action labels.

In general, our high-order relation reasoning block ACAR-Net is weakly-supervised, which only requires action labels as supervision.

{\flushleft \bf Instantiation.} We implement the High-order Relation Reasoning Operator $\text{HR}^2\text{O}$ as a location-wise attention operator, which is natural for modeling the connections between multiple first-order relations at the same spatial location. The operator $\text{HR}^2\text{O}_{\text{NL}}$ consists of one or two modified non-local blocks~\cite{wang2018non}. Since we are operating on a spatial grid of features, we replace the fully-connected layers in the non-local block with convolutional layers, and the attention vector is computed separately at every spatial location. Following \cite{wu2019long}, we also add layer normalization and dropout to our modified non-local block for improving regularization.

\begin{align*}
    Q^{i}&=\text{Conv2D}_{\text{q}}(F^{i})\\
    K^{j}&=\text{Conv2D}_{\text{k}}(F^{j})\\
    V^{j}&=\text{Conv2D}_{\text{v}}(F^{j})\\
    S^{i,j}_{x,y}&=\left\langle Q^{i}_{x,y},K^{j}_{x,y}\right\rangle/\sqrt{d}\\
    A^{i}_{x,y}&=\text{Softmax}_j(S^{i,j}_{x,y})\\
    W^{i}_{x,y}&=\text{ReLU}\left(\text{norm}\left(\sum_{j}A^{i,j}_{x,y}V^{j}_{x,y}\right)\right)\\
    H^{i}&=\text{dropout}(\text{Conv2D}_{\text{f}}(W^{i}))
\end{align*}

For saving memory, spatial $2\times2$ max pooling is applied by default to the first-order relation maps before feeding them into our operator. The high-order relation map $H^i$ will be spatially average-pooled, and then channel-wise concatenated to the basic actor RoI feature vector $A_i$ for final classification. All relation vectors are of dimension $d=512$ in our implementation.

\subsection{Actor-Context Feature Bank}
\label{sec:acfb}

Inspired by the Long-term Feature Bank (LFB) \cite{wu2019long}, which creates a feature bank over a large time span to facilitate first-order actor-actor relation reasoning, we consider creating an \textbf{Actor-Context Feature Bank} $F_{\text{bank}}$ which is built upon the first-relation features computed in our ACAR-Net. Formally, $F_{\text{bank}} = [F_{0}, F_{1}, \cdots, F_{T-1}]$, where $F_t$ is the first-order actor-context relation map extracted from a short video clip around time $t$. This bank of features can be obtained by running a trained ACAR-Net over the entire video at evenly spaced intervals (by default 1 second) and saving the intermediate first-order relation maps. Different from the original LFB, our relational feature preserves spatial context information. Equipped with such a relational feature bank, our ACAR-Net can leverage its High-order Relation Reasoning Operator for reasoning actor-context-actor relations over a much longer time span, and thus better capture what is happening in the entire video for achieving more accurate action localization at the current time stamp.

{\flushleft \bf Instantiation.} We experiment on ACFB with the aforementioned $\text{HR}^2\text{O}_{\text{NL}}$ implementation of high-order relation reasoning in ACAR-Net. We stack two modified non-local blocks, and replace the self-attention mechanism with an attention between current and long-term actor-context relations. For AVA videos, we set the long-term time span to $\sim$20 seconds, and for a Kinetics video, the bank simply spans across the entire video whose length is at most 10 seconds. For faster convergence, we do not apply spatial max pooling before $\text{HR}^2\text{O}_{\text{NL}}$.

%% file: 02-Experiments.tex
\section{Experiments}
\label{sec:Experiments}

\subsection{Implementation Details}

{\flushleft \bf Dataset.} 
For this year's challenge, Kinetics-700~\cite{carreira2019short} videos with AVA~\cite{gu2018ava} style annotations are introduced. The new AVA-Kinetics dataset~\cite{li2020ava} of spatio-temporally localized atomic visual actions contains over 238k unique videos and more than 624k annotated frames. For AVA, box annotations and their corresponding action labels are provided on key frames of 430 15-minute movie clips with a temporal stride of 1 second, while for Kinetics only a single frame is annotated for each video. Following the guidelines of the challenge, we evaluate on 60 action classes, and the performance metric is mean Average Precision (mAP) using a frame-level IoU threshold of 0.5.

{\flushleft \bf Person Detector.} As for person detector on key frames, we adopted the detection model from \cite{liu20201st}, which is a Faster R-CNN~\cite{ren2015faster} with an SENet-154-FPN-TSD \cite{hu2018squeeze,lin2017feature, song2020revisiting} backbone. The model is pre-trained on OpenImage~\cite{OpenImages}, and then fine-tuned on AVA and Kinetics respectively. The final models obtain 95.8 AP@50 on the AVA validation set, and 84.4 AP@50 on the Kinetics validation set.

\begin{table*}[!t]
\centering
\begin{tabular}{@{}l|x{40}|x{20}|x{40}|x{40}@{}}
 model & head & 3S+F & val mAP & test mAP\\
\shline
 {SlowFast, R101, $8\times8$} & Linear & \multirow{4}{*}{\XSolid} & 32.98 & - \\
 {SlowFast, R101, $8\times8$} & ACAR & & 34.58 & - \\
 {SlowFast, R152, $8\times8$} & ACAR & & 35.12 & 34.99 \\
 {SlowFast, R101, $8\times8$} & ACFB & & \textbf{35.84} & - \\ 
 \hline
 {SlowFast, R101, $8\times8$} & Linear & \multirow{5}{*}{\Checkmark}& 33.96 & - \\
 {SlowFast, R101, $8\times8$} & ACAR & & 35.44 & - \\
 {SlowFast, R152, $8\times8$} & ACAR & & 35.96 & - \\
 {SlowFast, R101, $8\times8$} & ACFB & & \underline{36.36} & - \\ 
 {SlowFast, ensemble} & mixed & & \textbf{40.49} & \textbf{39.62}
\end{tabular}
\caption{\textbf{AVA-Kinetics v1.0 results}. "3S+F" in the third column refers to inference with 3 scales and horizontal flips. Models submitted to the test server are trained on both training and validation data. \label{tab:results}}
\vspace{-3mm}
\end{table*}
\begin{table*}[!t]
\centering
\begin{tabular}{@{}l|x{40}|x{60}|x{60}|x{60}@{}}
 model & head & dataset & AVA val mAP & AVA test mAP \\
\shline
 {SlowFast, R101, $8\times8$} & Linear & AVA & 30.30 & - \\
 {SlowFast, R101, $8\times8$} & Linear & AVA-Kinetics & \textbf{32.25} & - \\
 \hline
 {SlowFast, R101, $8\times8$} & ACAR & AVA & 32.29 & - \\
 {SlowFast, R101, $8\times8$} & ACAR & AVA-Kinetics & \textbf{34.15} & - \\
 \hline
 {SlowFast, ensemble~\cite{feichtenhofer2019report}} & Linear & AVA & - & 34.25 \\
 {SlowFast, ensemble} & mixed & AVA-Kinetics & - & \textbf{38.30}
\end{tabular}
\caption{\textbf{Effect of adding Kinetics data on AVA v2.2}. The four single models are tested with single scale (256). \label{tab:kinetics}}
\vspace{-3mm}
\end{table*}

{\flushleft \bf Backbone Network.} We use SlowFast networks~\cite{feichtenhofer2019slowfast} as the backbone in our localization framework, and we also increase the spatial resolution of res$_5$ by $2\times$. We use SlowFast R101 and R152 instantiations with input sampling $T\times\tau=8\times8$ and $16\times8$ (without non-local) pre-trained on the Kinetics-700 dataset~\cite{carreira2019short}.

{\flushleft \bf Training.}  We use per-class binary cross entropy loss as the training loss function. Since one person should only have one pose label, following \cite{xia2019three}, we apply a softmax function instead of sigmoid to the logits corresponding to pose classes.

We train all models in an end-to-end fashion (except the feature bank part) using synchronous SGD with a minibatch size of 32 clips. We freeze batch normalization layers in the backbone network. We train most models for 55k steps (6 epochs on the training set) with a base learning rate of 0.064, which is decreased by a factor of 10 at iterations 51k and 53k. A few models are trained with an extended 8-epoch schedule for final ensemble. We perform linear warm-up~\cite{goyal2017accurate} during the first 9k iterations. For models submitted to the test server, we train on both training and validation data for the same number of epochs. We use weight decay of $10^{-7}$ and Nesterov momentum of 0.9. For a model with $T\times8$ input sampling, we use $4T$ frames centered at the key frame as input, sampled with a temporal stride of 2. Note that in some Kinetics data, the annotated timestamps are too close to the end of the videos, and in these cases we simply sample the last $4T$ frames. In order to better preserve spatial structure, we do not use spatial random cropping augmentation. Instead, we only scale the shorter side of the input frames to 256 pixels, and zero pad the longer side to the same size in order to simplify mini-batch training. For AVA, we use both ground-truth boxes and predicted human boxes from \cite{wu2019long} for training, and only ground-truth boxes for generating feature banks. For Kinetics, we only use ground-truth boxes for training, and our detection boxes for generating feature banks. The bank of features are extracted from short clips sampled with a temporal stride of 1 second from both AVA and Kinetics videos. We use bounding box jittering augmentation, which randomly perturbs box coordinates by a scale at most 7.5\% relative to the original size of the bounding box during training.

{\flushleft \bf Inference.} At test time, we use AVA detections with confidence $\ge 0.7$ and Kinetics detections with confidence $\ge 0.65$. We scale the shorter side of input frames to 256 pixels, and apply the backbone feature extractor fully-convolutionally. We also report results tested with three spatial scales $\{256,288,320\}$ and horizontal flips.

\subsection{Main Results}

We present our results on AVA-Kinetics v1.0 in Table~\ref{tab:results}. The default backbone instantiation is SlowFast R101 $8\times8$. The simplest baseline, linear classifier head, already has nearly 33mAP. Switching to our ACAR-Net still brings about a significant 1.6 increase in mAP. This highlights the importance of modeling high-order relations. Further adding long-term support (ACFB head) gives a total boost of 2.86mAP. We also experiment on a more advanced backbone (SlowFast R152 $8\times8$) which brings some extra improvement in performance (+0.54mAP). We re-trained this SlowFast R152 model on training and validation data, and submitted it to the test server. Its performance almost did not drop (-0.13mAP).

For final ensemble, we combined predictions of 20 models with 4 different backbones, several different heads (Linear, LFB~\cite{wu2019long}, ACAR, ACFB), and different schedules (6 \textit{vs.} 8 epochs). Similar to \cite{akiba2018pfdet}, for each action class, we set weights in the ensemble according to APs of the models on this class.

\subsection{Ablation Experiments}

{\flushleft \bf Effect of Adding Kinetics Data.} For two SlowFast R101 $8\times8$ models, we train the same model with two different datasets, AVA-Kinetics and AVA only, and evaluate on AVA v2.2 validation set. As shown in Table~\ref{tab:kinetics}, adding Kinetics data brings consistent mAP increases (roughly +2mAP) to these two models. Moreover, with the help of both high-order relation reasoning and Kinetics data, our ensemble achieves a significant enhancement of \textbf{+4.05mAP} on AVA v2.2 test set compared to last year's winner~\cite{feichtenhofer2019report}.

{\flushleft \bf Different Detectors.} We investigate the effect of person detection AP@50 on action detection mAP. We perform the comparison on SlowFast R101 $8\times8$ with ACAR head. As presented in Table~\ref{tab:avadet} and \ref{tab:kineticsdet}, person detection AP@50 on Kinetics is much lower than that on AVA. In addition, even though our detector have reached 95.8 AP@50 on AVA person detection, there is still a large gap (8.1mAP) in final mAP between our detection and ground-truth (GT). These results suggest that how to improve person detection for action localization still remains to be explored.
\begin{table}[!h]
\centering
\begin{tabular}{@{}l|x{30}|x{25}@{}}
 detector & AP@50 & mAP\\
\shline
 LFB~\cite{wu2019long} & 93.9 & 33.73 \\
 Ours & \textbf{95.8} & \underline{34.15} \\
 GT & - & \textbf{42.25}
\end{tabular}
\caption{\textbf{Different detectors on AVA v2.2}. Final action detection results are evaluated on SlowFast R101 $8\times8$ with ACAR head. \label{tab:avadet}}
\vspace{-3mm}
\end{table}
\begin{table}[!h]
\centering
\begin{tabular}{@{}l|x{30}|x{25}@{}}
 detector & AP@50 & mAP\\
\shline
 Ours (AVA) & 77.2 & 28.41 \\
 Ours & \textbf{84.4} & \underline{30.88} \\
 GT & - & \textbf{43.60}
\end{tabular}
\caption{\textbf{Different detectors on Kinetics v1.0}. We test our AVA detector on Kinetics data with confidence threshold set to 0.9. Final action detection results are evaluated on SlowFast R101 $8\times8$ with ACAR head.  \label{tab:kineticsdet}}
\vspace{-3mm}
\end{table}

\subsection{Pre-training on Kinetics}

We used 4 SlowFast models pre-trained from scratch on Kinetics-700 classification task. We show their single center crop ($224\times224$) accuracy on Kinetics-700 validation set in Table~\ref{tab:pretrain}. Note that our models might have not reached full convergence due to time limitation.
\begin{table}[!h]
\centering
\begin{tabular}{@{}l|x{30}@{}}
 model & val acc.\\
\shline
 {SlowFast, R101, $8\times8$} & 60.2 \\
 {SlowFast, R101, $16\times8$} & 62.6 \\
 {SlowFast, R152, $8\times8$} & 63.4 \\
 {SlowFast, R152, $16\times8$} & 66.1
\end{tabular}
\caption{\textbf{Pre-training on Kinetics-700}. All results are obtained with spatial and temporal center crop (single crop). \label{tab:pretrain}}
\vspace{-3mm}
\end{table}